\documentclass[10pt,twocolumn,letterpaper,final]{cvpr}

\usepackage{times}
\usepackage{epsfig}
\usepackage{graphicx}
\usepackage{amsmath}
\usepackage{amssymb}
\usepackage{graphicx}
\usepackage{float}
\usepackage{subfigure}
\usepackage{amsmath}
\usepackage{colortbl}
\usepackage{url}

\usepackage{cite}
\usepackage{ftnright}
\usepackage{lipsum}
\makeatletter

\newcommand{\printfnsymbol}[1]{%
	\textsuperscript{\@fnsymbol{#1}}%
}
\newcommand\blfootnote[1]{%
\begingroup
\renewcommand\thefootnote{}\footnote{#1}%
\addtocounter{footnote}{-1}%
\endgroup
}
\makeatletter
\@addtoreset{section}{part}
\makeatother

\makeatother

\renewcommand\footnoterule{\hrule width 2.5cm}

\usepackage[pagebackref=true,breaklinks=true,colorlinks,bookmarks=false]{hyperref}

\def\cvprPaperID{2146} 
\def\confYear{CVPR 2021}

\begin{document}

\title{Learning to Fuse Asymmetric Feature Maps in Siamese Trackers}

\author{
Wencheng Han$^{*1}$, Xingping Dong$^{*2}$, Fahad Shahbaz Khan$^3$, Ling Shao$^{2}$, Jianbing Shen$^{\dagger2,1}$\\

$^1$Beijing Institute of Technology, $^2$Inception Institute of Artificial Intelligence \\ $^3$Mohamed Bin Zayed University of Artificial Intelligence,UAE \\
{\tt\small wencheng@bit.edu.cn}, {\tt\small xingping.dong@gmail.com}, {\tt\small fahad.khan@liu.se} \\ {\tt\small ling.shao@ieee.org}, {\tt\small shenjianbingcg@gmail.com}
}




\pagestyle{empty}
\maketitle

\thispagestyle{empty}
\blfootnote{$*$Equal contribution. $\dagger$ Corresponding author.}
\blfootnote{Our codes are available at: \url{https://github.com/wencheng256/SiamBAN-ACM}}
\begin{abstract}
Recently, Siamese-based trackers have achieved promising performance in visual tracking. Most recent Siamese-based trackers typically employ a depth-wise cross-correlation (DW-XCorr) to obtain multi-channel correlation information from the two feature maps (target and search region). However, DW-XCorr has several limitations within Siamese-based tracking: it can easily be fooled by distractors, has fewer activated channels and provides weak discrimination of object boundaries. Further, DW-XCorr is a handcrafted parameter-free module and cannot fully benefit from offline learning on large-scale data.

We propose a learnable module, called the asymmetric convolution (ACM), which learns to better capture the semantic correlation information in offline training on large-scale data. Different from DW-XCorr and its predecessor (XCorr), which regard a single feature map as the convolution kernel, our ACM decomposes the convolution operation on a  
concatenated feature map into two mathematically equivalent operations, thereby avoiding the need for the feature maps  to  be  of the same  size (width and height) during concatenation. Our ACM can incorporate useful prior information, such as bounding-box size, with standard visual features. Furthermore, ACM can easily be integrated into existing Siamese trackers based on DW-XCorr or XCorr. To demonstrate its generalization ability, we integrate ACM into three representative trackers: SiamFC, SiamRPN++ and SiamBAN. Our experiments reveal the benefits of the proposed ACM, which outperforms existing methods on six tracking benchmarks. On the LaSOT test set, our ACM-based tracker obtains a significant improvement of 5.8\% in terms of success (AUC), over the baseline.

\end{abstract}

\section{Introduction}
\label{section:introudtion}
Visual tracking is a challenging problem, where the task is to estimate the state of an arbitrary target in each frame of a video, given only its location in the initial frame. Recently, trackers based on Siamese networks have gained attention due to their combined advantage of high speed and tracking performance. The pioneering method, SiamFC~\cite{bertinetto2016fully}, utilizes Siamese networks to extract deep convolutional features from the template in the initial frame of a video and instances inside the search regions of other frames. A cross correlation layer (XCorr) is then used to compute the similarity between the template and instances. Consequently, the instance with the highest similarity score is considered the target. The XCorr in SiamFC produces a single-channel response map and assumes the
target is located near the highest response. As an extension, SiamRPN~\cite{li2018high} formulates the tracking problem as one-shot detection. It introduces a region proposal network (RPN)~\cite{ren2015faster} and utilizes up-channel cross correlation (UP-XCorr). However, UP-XCorr imbalances the parameter distribution, making the training optimization hard. To address this issue, SiamRPN++ introduces a depth-wise correlation
(DW-XCorr) to efficiently generate a multi-channel correlation feature map. Due to its efficiency, several recent Siamese trackers~\cite{guo2020siamcar,chen2020siamese,yu2020deformable,xu2020siamfc++,du2020correlation} also employ DW-XCorr in their frameworks.

\begin{figure}
  \centering
  \includegraphics[width = 8cm]{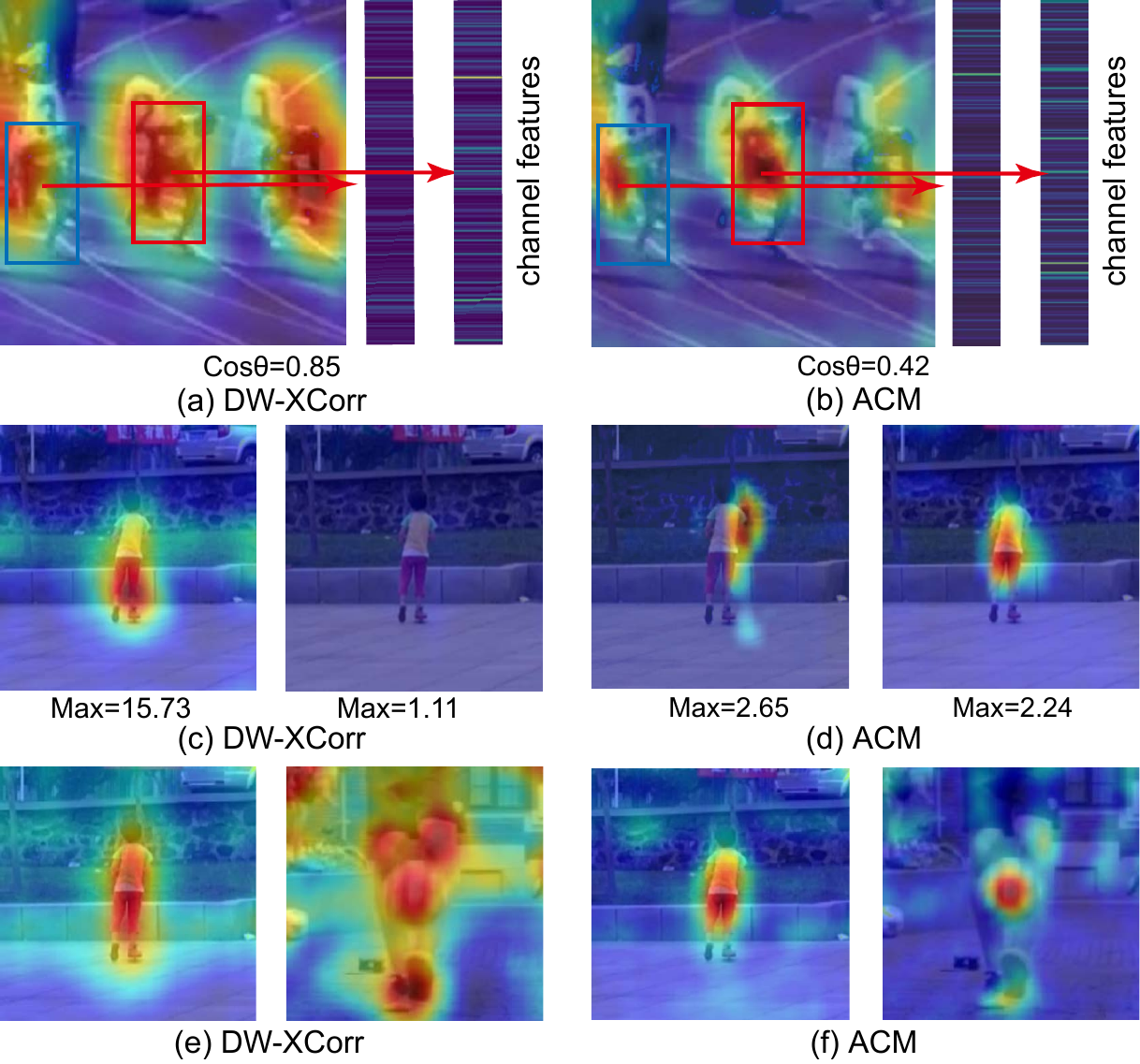}
  \caption{
  \textbf{Comparison between DW-XCorr and ACM in terms of being fooled by distractors (first row), information distribution across channels (second row) and background suppression to better discriminative target boundaries (third row).}
   DW-XCorr produces similar responses for distractors and the target  (Fig.~\ref{Fig:1}\textbf{a}). In contrast, ACM produces more distinct responses (Fig.~\ref{Fig:1}\textbf{b}). In both cases (a and b), red arrows indicate the feature vectors extracted from the correlation feature maps of the corresponding pixels, followed by computing the cosine similarity ($cos\theta = \frac{A\cdot B}{\left\|A\right\| \left\|B\right\|}$) between the two feature vectors ($A$ and $B$). Only a few channels of DW-XCorr have high response when tracking a desired target (Fig.~\ref{Fig:1}\textbf{c}). Instead, more channels of ACM map carries high response with different semantic information, such as top right corner (left) or center of target (right), as shown in Fig.~\ref{Fig:1}\textbf{d}. We show two example feature channels for DW-XCorr and ACM. DW-XCorr maps are blurry and do not accurately capture shape of target (Fig.~\ref{Fig:1}\textbf{e}). In comparison, AC maps suppresses the background, providing clear boundaries of the target (Fig.~\ref{Fig:1}\textbf{f}).
  }\label{Fig:1}
\end{figure}


As discussed above, most recent Siamese trackers employ DW-XCorr to compute the similarity between the template and instances. However, both DW-XCorr and its predecessor XCorr are handcrafted parameter-free modules and are not able to fully benefit from large-scale offline learning. DW-XCorr has several limitations in the context of tracking. First, it produces similar correlation responses for the target and distractors of homogeneous appearance. To demonstrate this, we analyze the similarity between  DW-XCorr features of a target and its distractors in Fig.~\ref{Fig:1}a. The heatmap is generated by performing an L1 normalization ($\left\|\mathbf{x}\right\|_1=\sum_{i=1}^{n}|x_i|$, where $x$ is a pixel in the correlation feature map and $n$ is the number of channels) on every pixel in the DW-XCorr features.
As can be seen, DW-XCorr produces high responses (\textit{i.e.} feature norms) not only near the target (the red rectangle), but also near other instances. We compute the cosine similarity between the target and one distractor (the green rectangle) and observe a high value ($cos\theta$ $>$ 0.8), indicating that DW-XCorr produces similar results for both. This makes it difficult for RPN to effectively discriminate the desired target from distractors.

The second limitation is that only a few channels in the DW-XCorr feature map are activated, \textit{i.e.} have a high response when tracking a particular target~\cite{li2019siamrpn}. To perform cross-correlation, features of different targets are desired to be orthogonal and distributed in different channels, so that correlation feature channels of different targets are suppressed and only a few channels of the same target are activated. The suppressed channels are unable to help RPN in making robust and precise predictions and can reduce the capacity of the model. As shown in Fig.~\ref{Fig:1}c, the maximum value of a channel with middle response is significantly lower than the global maximum value. This indicates that these channels contribute little to the final predictions. Last, DW-XCorr often produces responses at irrelevant background. As a consequence, correlation maps are often blurry and do not have clear boundaries, as shown in Fig.~\ref{Fig:1}e. This is likely to hinder RPN from making accurate and robust predictions.

The aforementioned shortcomings of DW-XCorr and its predecessor XCorr, within Siamese-based trackers, motivate us to look into designing a new module that learns to fuse feature maps by benefiting from offline learning on large-scale data. In case of two feature maps (\eg. the template and sub-window in a search image) having the same size, a straightforward way is to concatenate (fuse) them and then learn a method for joint training by adding convolutional layers. Here, the additional convolutional layers can learn to discriminate the target and background. However, such a concatenation strategy is non-trivial in the case of Siamese-based trackers since the two feature maps are of different sizes (height and width). Further, the concatenation of feature maps of different sizes is desired to be performed in an efficient manner to meet the real-time requirements during inference. 



\subsection{Contributions}
\label{section:contributions}

We introduce a novel module, called the \textit{asymmetric convolution (ACM)}, that avoids the need for the feature maps to be of the same size during concatenation. Our ACM decomposes the convolution operation on a concatenated feature map into two mathematically equivalent operations. First, it performs convolutions on two feature maps independently using kernels of the same size as that of the template feature map. Then, it performs a summation on the resulting feature maps, through broadcasting~\cite{numpy}. By utilizing the broadcasting of matrix addition, we efficiently compute the summation on these different-sized feature maps.

The proposed ACM produces more discriminative features, as shown in Fig.~\ref{Fig:1}b, with respect to the target (the red rectangle) and distractors (the green rectangle). This enables the tracker to make more robust predictions. Further, the maximum values of different channels in our ACM are closer, which indicates that more channels carry useful information, as shown in Fig.~\ref{Fig:1}d. At the same time, ACM can effectively suppress background, thereby providing clear boundaries for the target, as in Fig.~\ref{Fig:1}f. We validate these advantages by conducting an extensive analysis on 50k different image pairs from the LaSOT train set~\cite{fan2019lasot}. Details are presented in \S\ref{section:ac}.


In addition to overcoming the aforementioned limitations of DW-XCorr, the proposed ACM is flexible and can also incorporate useful additional information. Here, we incorporate prior information in the form of bounding-box (b-box) size (height and width) from the initial frame in a video. This prior information helps to overcome the lack of accurate target-box locations in the template image, thereby providing guidance to the RPN heads.  
Furthermore, we show the generalization ability by replacing the standard DW-XCorr or XCorr with our ACM in three representative Siamese-based trackers: SiamFC~\cite{bertinetto2016fully}, SiamRPN++~\cite{li2019siamrpn} and SiamBAN~\cite{chen2020siamese}.
Comprehensive experiments on six tracking benchmarks show the benefits of our ACM, leading to favorable performance against existing methods. On the large-scale LaSOT test set~\cite{fan2019lasot}, our ACM-based trackers (SiamFC-ACM, SiamRPN++ACM and SiamBAN-ACM) achieve relative gains of 8.6\%, 5.7\%  and 11.3\%, in terms of area-under-the-curve (AUC), over their respective baselines (SiamFC, SiamRPN++ and SiamBAN).

\section{Related Work}

Recently, deep learning has pervaded computer vision with great success in a variety of tasks, including object tracking~\cite{wang2013learning,bertinetto2016learning,nam2016learning,tao2016siamese,bertinetto2016fully,held2016learning,yun2017action,kosiorek2017hierarchical,song2018vital,park2018meta,pu2018deep,danelljan2019atom}.
Several deep learning-based trackers learn a classifier online to distinguish the target from the background and distractors~\cite{wang2013learning,nam2016learning,song2017crest,sun2018correlation,danelljan2019atom}. 
The MDNet~\cite{nam2016learning} tracker employs a CNN trained offline from multiple annotated videos. During evaluation, it learns a domain-specific detector online to discriminate between the background and foreground. ATOM~\cite{danelljan2019atom} comprises two dedicated components: target estimation, which is trained offline, and classification trained online. DiMP~\cite{bhat2019learning} employs a meta-learning based architecture, trained offline, that predicts the weights of the target model. The recently introduced KYS~\cite{bhat2020know} extends DiMP by exploiting scene information to improve the results. 




Several existing deep trackers \cite{bertinetto2016fully,li2018high,li2019siamrpn,chen2020siamese,guo2020siamcar,xu2020siamfc++} are based on Siamese networks and focus on learning a universal discriminator during large-scale offline learning.
These trackers formulate the task as a general similarity computation between the target template and the search region. The pioneering work, SiamFC~\cite{bertinetto2016fully}, introduced the XCorr layer to combine feature maps and can run at a speed of 100 frames per second (FPS).  
Since this work, several researchers have tried to further mine the potentiality of the Siamese framework by designing different Siamese architectures~\cite{he2018twofold,wang2018learning,zhang2018structured,dong2019quadruplet}, using a powerful training loss~\cite{dong2018triplet}, learning efficient Siamese networks~\cite{liu2019teacher-students}, learning a dynamic network~\cite{guo2017learning}, utilizing deep reinforcement learning~\cite{huang2017learning,dong2018hyperparameter,dong2019dynamical}, and so on~\cite{wang2018sint,yang2018learning,shen2019visual,liang2020local}. 
SiamRPN++~\cite{li2019siamrpn} and SiamDW~\cite{zhang2019deeper} overcome the issues of previous Siamese-based trackers that restrict them to using only relatively shallow networks. Specifically, they address the problems stemming from destroying the strict translation invariance and introduce modern deep networks, such as, ResNet~\cite{he2016deep}, and ResNeXt~\cite{xie2017aggregated}, into Siamese trackers. SiamRPN++ utilizes a depth-wise  correlation (DW-XCorr) to efficiently generate a multi-channel correlation feature map. The recently introduced SiamBAN\cite{chen2020siamese} and SiamCAR\cite{guo2020siamcar} also employ DW-XCorr and use an anchor-free strategy to predict bounding-boxes (b-boxes) directly without pre-defined anchor boxes.

\noindent \textbf{Our Approach:} As discussed earlier, most recent Siamese trackers typically employ a handcrafted module, DW-XCorr, to compute the similarity between the template and instances. Both DW-XCorr and its predecessor XCorr are not able to fully benefit from large-scale offline learning and have several limitations, including being easily fooled by distractors and providing weak discrimination of the object boundaries. To address these issues, we propose a new module (ACM) that learns to better capture semantic information from large-scale data during offline training. Our ACM produces more discriminative features with respect to the target and distractors, contains more activated channels carrying useful information and effectively suppresses the background, thereby providing clear boundaries of the target. Furthermore, our ACM is flexible and generic, enabling easy integration into existing Siamese trackers. We show this by integrating our ACM into three Siamese trackers and demonstrate its effectiveness on six benchmarks.

\section{Method}



\subsection{Siamese Networks for Tracking}
Siamese networks formulate the tracking task as learning a general similarity map between the feature maps extracted from the target template and the search region. When certain sliding windows in the search region are similar to the template, responses in these windows are high~\cite{bertinetto2016fully}.
These networks are designed as Y-shaped, with two branches: one for the template $\mathbf{z}$ and the other for the search region $\mathbf{x}$.
The two branches share the same network $\varphi$ with parameters $\theta$ and produce two feature maps $\mathbf{\bar{z}}=\varphi(\mathbf{z};\theta) \in \mathbb{R}^{C \times \eta \times \omega }$ and $\mathbf{\bar{x}} =\varphi(\mathbf{x};\theta)  \in \mathbb{R}^{C \times H \times W}$. These two feature maps have the same channel number ($C$) but different sizes ($\eta \times \omega$ {\it vs.} $H \times W$), where $\eta \leq H$ and $\omega \leq W$. Then, a function $f$ is used to combine the feature maps and generate a similarity map $\mathbf{c} \in \mathbb{R}^{1 \times (H - \eta + 1) \times (W - \omega + 1)}$, where the center of the target is most likely found at the position with the highest response. Usually, $f$ is an XCorr operation $*$ between $\mathbf{\bar{x}}$ and $\mathbf{\bar{z}}$. The formulation is as follows:
\begin{equation}\label{eq:xcorr}
    \mathbf{c} = f(\mathbf{\bar{z}}, \mathbf{\bar{x}}) = \varphi(\mathbf{z};\theta) * \varphi(\mathbf{x};\theta).
\end{equation}

To further improve the performance of Siamese-based trackers, SiamRPN~\cite{li2018high} adds region proposal network (RPN)~\cite{ren2015faster} to generate bounding-boxes (b-boxes) for each frame of a tracking sequence.The RPN contains two XCorr modules to extract correlation maps and two heads on them to perform anchor classification and regression, respectively. This is different to previous Siamese trackers, such as SiamFC, where the b-box is not explicitly regressed and is typically set based on the size that best matches the search region. While SiamRPN utilizes an RPN, it employs up-channel cross correlation (UP-XCorr), which imbalances the parameter distribution, making the training optimization difficult. SiamRPN++~\cite{li2019siamrpn} addresses this issue by introducing a depth-wise correlation (DW-XCorr) to efficiently generate a multi-channel correlation feature map $\mathbf{c}_{dw}$, as in Fig.~\ref{Fig:AC}a. The formulation is as follows:

\begin{equation}
\begin{aligned}
      &\mathbf{c}_{dw} = f( \mathbf{\bar{z}}, \mathbf{\bar{x}}) = \mathbf{\bar{z}} \otimes \mathbf{\bar{x}}; \\
      &\mathbf{c}_{dw}\in \mathbb{R}^{N \times (H - \eta + 1) \times (W - \omega + 1),}
\end{aligned}
\end{equation}
where $\otimes$ is a depth-wise convolution~\cite{howard2017mobilenets} of two feature maps, and $N$ is the number of channels. Then, 
the features are fed into the RPN heads to produce the final tracking b-box. The RPN heads are usually constructed with sequences of $1 \times 1$ convolutional layers, including the classification module $\mathcal{H}_{cls}$, which predicts the classification score of each b-box candidate, and the regression module $\mathcal{H}_{loc}$, which obtains the details (size in terms of width and height) of each b-box. By applying these heads to the correlation maps, we can obtain the score map $\mathbf{m}_{cls} \in \mathbb{R}^{2 \times A \times (H - \eta + 1) \times (W - \omega + 1)}$ and b-box map $\mathbf{m}_{loc} \in \mathbb{R}^{4 \times A \times (H - \eta + 1) \times (W - \omega + 1)}$:
\begin{equation}
    \mathbf{m}_{cls} = \mathcal{H}_{cls}(\mathbf{c}_{dw}^{cls};\theta_{cls}),~~~~~
    \mathbf{m}_{loc} = \mathcal{H}_{cls}(\mathbf{c}_{dw}^{loc};\theta_{loc}).
\end{equation}
As we can see, the fusion method $f$ is crucial for Siamese-based trackers. However, both the XCorr and depth-wise XCorr (DW-XCorr) are parameter-free methods and therefore cannot fully benefit from large-scale training. Further, they have several limitations as described in \S\ref{section:introudtion}. Our asymmetric convolution module (ACM) addresses these limitations by introducing an asymmetric convolution (AC) as $f( \mathbf{\bar{z}}, \mathbf{\bar{x}}; \theta_{ac})$. With the parameter $\theta_{ac}$, AC can be optimized during training and finds the a better way to fuse  $\mathbf{\bar{z}}$ and $\mathbf{\bar{x}}$.



\subsection{Asymmetric Convolution}\label{section:ac}
\begin{figure}
  \centering
  \includegraphics[width = 0.45\textwidth]{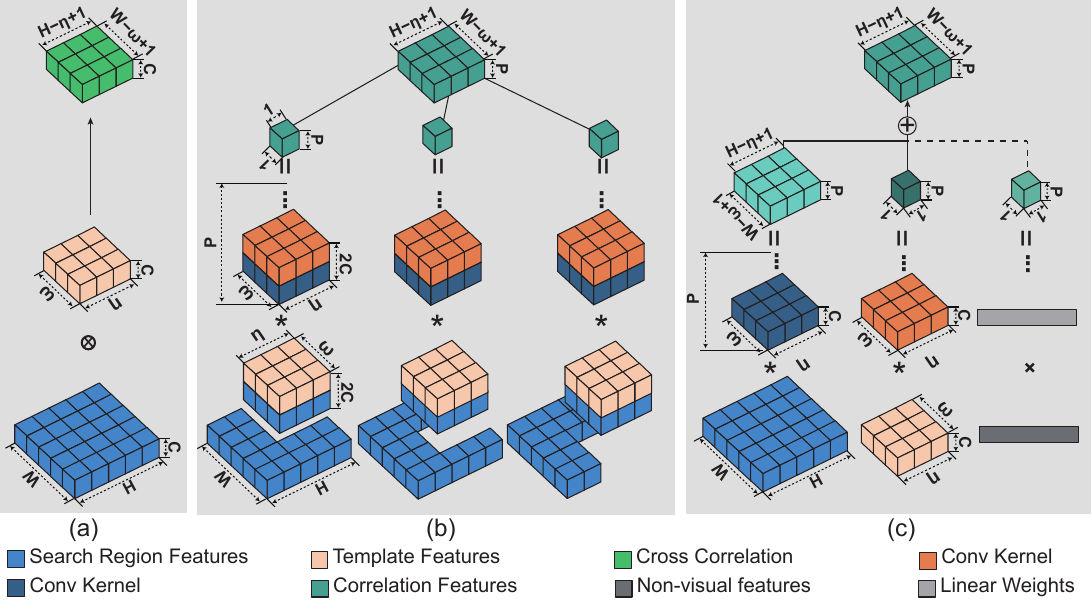}
  \caption{ \textbf{Comparison of \textbf{(c)} AC with \textbf{(a)} DW-XCorr and \textbf{(b)} a naive strategy to fuse different-sized feature maps.} \textbf{(a):} DW-XCorr uses a $C$ channel feature map extracted from the template as kernel and convolves instance feature maps in a depth-wise manner to generate a $C$ channel correlation feature map. \textbf{(b):} A naive strategy to perform concatenation on different-sized feature maps (template and search region) is to first split the search region feature map into $n$ sub-windows of the same size as that of the template feature map. Then, $n$ different sub-windows and the template are concatenated along channel axis,
followed by a convolution  to generate a new feature during offline training. \textbf{(c):} AC efficiently concatenates different-sized feature maps by first separately convolving the two feature maps (template and search) using kernels of same size as that of the template feature
map. Then, it computes
summation on these different-sized feature maps through broadcasting. In addition, our AC possesses the ability to incorporate useful non-visual features (dashed line), such as b-box size. }
  \label{Fig:AC}
\end{figure}

 
Different from handcrafted methods (\eg, DW-XCorr and XCorr) for fusing features in Siamese networks, we look into how to concatenate two different-sized feature maps and learn a fusion during offline training on large-scale data. Learning to fuse feature maps during offline training is expected to provide rich prior information, enabling the fusion method to better adapt to different challenging situations, such as motion blur, deformation, fast motion and clutter. However, an efficient direct concatenation of these feature maps is challenging due to the different sizes of the template and search image. To this end, we investigate the problem of efficiently fusing feature maps of different sizes. A straightforward way (Fig.~\ref{Fig:AC}b) is to first split the the search region feature map into $n$ sub-windows of the same size as that of the template feature map. It is woth noting that every sub-window is a sliding window here. Then, $n$ different sub-windows and the template can be concatenated along the channel axis, followed by a convolution operation to produce a new feature $\mathbf{v}_i$. However, such a strategy (Fig.~\ref{Fig:AC}b) based on direct convolution on the concatenated feature map is computationally expensive, since the convolution operation is required to be repeated for each sub-window.
To circumvent this problem, we introduce a mathematically equivalent procedure, called the asymmetric convolution (AC), that replaces this direct convolution on the concatenated feature map with two independent convolutions (Fig.~\ref{Fig:AC}c). For a sub-window $n$, our AC,  comprising two independent convolutions followed by a summation, is mathematically equivalent to the direct convolution on the concatenated feature map:



\begin{equation}
\label{eq:v_i}
\begin{aligned}
\mathbf{v}_i =
 \left[
 \begin{matrix}
  \theta_z  & \theta_x
 \end{matrix} 
  \right] *
 \left[
 \begin{matrix}
 \mathbf{~\bar{z}}~ \\ ~\mathbf{\bar{x}}_i
 \end{matrix} 
  \right] 
  =  \theta_z * \mathbf{\bar{z}}  + \theta_x * \mathbf{\bar{x}}_i ; \\
  \mathbf{\bar{x}}_i \in \mathbb{R}^{C \times \eta \times \omega }, \theta_z, \theta_x \in \mathbb{R}^{P \times C \times \eta \times \omega }, \mathbf{v}_i \in \mathbb{R}^{P \times 1 \times 1 },
  \end{aligned}
\end{equation}
where $\mathbf{\bar{x}}_i$ is a window of $\mathbf{\bar{x}}$, $\theta_z$ is the kernel applied to $\mathbf{\bar{z}}$, and $\theta_x$ is that applied to $\mathbf{\bar{x}}$. After the convolution operation, the result $\mathbf{v}_i$ has a shape of $P \times 1 \times 1$. The left side of Eq. \ref{eq:v_i} is a convolution on a concatenated feature map of $\mathbf{\bar{z}}$ and $\mathbf{\bar{x}}_i$, and it is equivalent to the right side, {\it i.e.}, two independent convolutions and a summation. Next, we collect the features of all windows inside $\mathbf{\bar{x}}$ to formulate a new feature map $\mathbf{v}$, as follows:
\begin{equation}
\begin{aligned}
\mathbf{v} & =\{\mathbf{v}_i~|~i\in[1, n]\} \\
& = \{\theta_z * \mathbf{\bar{z}} + \theta_x  * \mathbf{\bar{x}_i} ~|~ i\in[1, n]\} \\
& = \theta_z * \mathbf{\bar{z}}   +_b \theta_x  * \mathbf{\bar{x},}
\end{aligned}
\end{equation}
where $+_b$ is a summation with broadcasting. We utilize the broadcasting method since it efficiently conducts arithmetic operations on matrices with different shapes and is widely available in scientific computing packages, including Numpy~\cite{numpy} and Pytorch~\cite{steiner2019pytorch}. 
{Through broadcasting, engines allow the dimensions of arrays to differ. Specifically, arrays with smaller sizes are virtually duplicated (that is, without copying any data in the memory and thus introducing little computational burden), so that the shapes of the operands match\cite{numpy}}.
Moreover, all sub-windows inside $\mathbf{\bar{x}}$ share the same convolution. Therefore, we replace $\{\theta_x * \mathbf{\bar{x}}_i |i\in[1, n]\}$ with $\theta_x * \mathbf{\bar{x}}$ for simplicity. In this way, we perform a convolution operation on two feature maps with different shapes, simultaneously. 
After applying a ReLU activation function, we obtain a new fusion $f( \mathbf{\bar{z}}, \mathbf{\bar{x}}; \theta_{ac})$  which can be optimized during training:
\begin{equation}
\begin{aligned}
\mathbf{c}_{ac} &= f( \mathbf{\bar{z}}, \mathbf{\bar{x}}; \theta_{ac})  \\
&= ReLU( \theta_z * \mathbf{\bar{z}}  +_b \theta_x * \mathbf{\bar{x}} ); \\
&\mathbf{c}_{ac} \in \mathbb{R}^{C \times (H - \eta + 1) \times (W - \omega + 1)}.
\end{aligned}
\end{equation}

\begin{figure}
  \centering
  \includegraphics[width = 8cm]{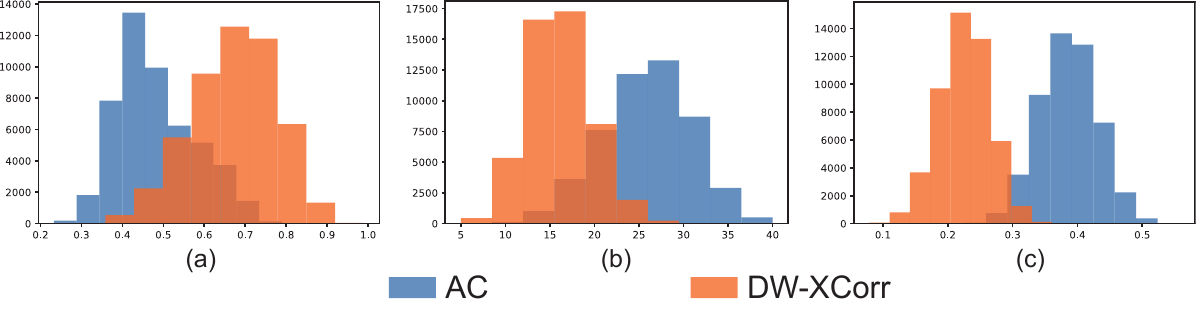}
  \caption{\textbf{Comparison between DW-XCorr and AC in terms of (a and b) producing more discriminative features for targets and distractors to avoid being fooled by the distractors and (c) 
  information distribution across correlation channels.} Comparison is performed on 50K different image pairs from LaSOT train set. \textbf{(a):} Cosine similarity between targets and distractors based on DW-XCorr and AC feature
maps, respectively. \textbf{(b):} Same as (a), except cosine similarity is replaced by Euclidean distance. In (b), the correlation feature maps are first normalized between [0,1] and then the Euclidean distance is computed between targets and distractors. \textbf{(c):} Average values over all maximum feature values of channels for DW-XCorr and AC, respectively. In each case, the maximum feature values are obtained by first performing a normalization (dividing the values by their global maximum value).
}\label{Fig:2}
\end{figure}
  
As discussed earlier, our AC benefits from the offline training and alleviates the limitations of DW-XCorr. To demonstrate that AC produces more discriminative features for the targets and distractors than XCorr, we perform an experiment in which we compute the cosine similarity between targets and distractors based on the AC and XCorr feature maps, respectively on 50k different image pairs from the LaSOT dataset. We set the target to be at the center of the search region and select the features located at the center of the AC and DW-XCorr maps to represent it. Then, we find the maximum response outside the b-box region and select features at this point to represent the distractor. Finally, the cosine similarity between the target and distractor features is computed to evaluate the discriminative ability of AC and DW-XCorr. Fig.~\ref{Fig:2}a shows that AC maps are less affected by distractors, producing more discriminative features, compared to DW-XCorr. Fig.~\ref{Fig:2}b shows a similar comparison but from a different perspective, where cosine similarity is replaced with the euclidean distance. Here, the correlation feature maps are first normalized between [0,1] and then the Euclidean distance is computed between targets and distractors. Further, AC maps contain more semantic information than DW-XCorr, as shown earlier in Fig.~\ref{Fig:1}b. We also validate, on same 50k image pairs from LaSOT, that AC channels provide more diversity when extracting correlation information, compared to DW-XCorr. We first normalize AC and DW-XCorr by dividing them by their global maximum value, and then calculate maximum values of each channel. Finally, average values over all channels are used to draw a comparison, shown in Fig.~\ref{Fig:2}c. Lastly, AC maps suppresses influence of irrelevant background better, compared to DW-XCorr, as shown earlier in Fig.~\ref{Fig:1}f. This helps RPN heads to more accurately predict the b-boxes.

\begin{figure}
  \centering
  \includegraphics[width = 0.45\textwidth]{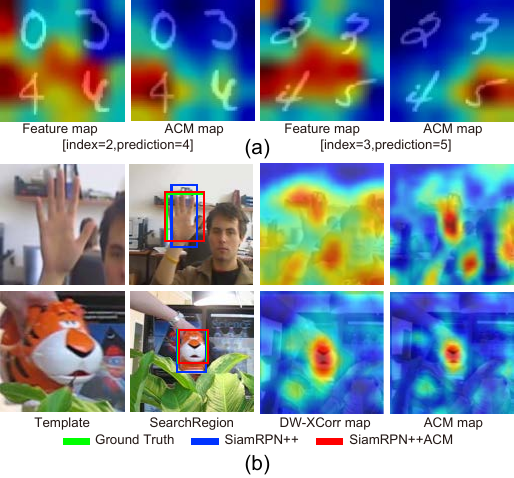}
  \caption{\textbf{(a): Effectiveness of our ACM} in fusing additional information (single number to indicate digit location) with visual feature maps for the task of digit prediction on MNIST. Here, "index" means the position to predict, and "prediction" is the predicted digit at this position. Indexes are 0,1 in the first row and 2,3 in the second row of the $2 \times 2$ matrix. The colors, superimposed on the images, are responses of feature maps where high responses are represented by warm colors.  \textbf{(b): Tracking comparison between our ACM-based tracker (SiamRPN++ACM) and the baseline (SiamRPN++)} on example frames, where the target is only part of an object (\eg, part of hand or body). Here,
  we also show DW-XCorr and ACM feature maps of the baseline and SiamRPN++ACM, respectively. Each feature map shown is obtained by taking the L1 norm of each pixel in the respective feature map. Our ACM map is able to focus on regions belonging to the target. Moreover, the integration of non-visual 1D b-box size features provides useful prior information to the RPN heads, leading to more accurate predictions.   
 }
  \label{Fig:index_exp}
\end{figure}
  


\subsection{Incorporating Prior Non-Visual Information}  \label{section:bbox}

As discussed earlier, our ACM is flexible and can incorporate additional (non-visual) information. Here, we show the integration of prior information in the form of target b-box size (width and height) from the initial frame. It is worth noting that traditional RPN head has no exact prior information about the target b-box which can be of arbitrary shape. ACM can provide such additional prior information, in terms of a b-box size, to the RPN head for accurate target localization. However, a b-box size is a one-dimensional feature and cannot be fed directly into 2D convolutional networks. Here, we regard a b-box size as a specific image feature with a size of $C_b \times 1 \times 1$, where $C_b$ is the channel number. In this way, we utilize ACM to fuse useful prior information, such as b-box size, with standard high-dimensional visual features representing template and search regions. 




We use the b-box size information from the initial frame in our tracking framework to distinguish features belonging to the target and provide guidance to the RPN heads:
\begin{equation}
\begin{aligned}
\mathbf{c}_{ac} &= f( \mathbf{\bar{z}}, \mathbf{\bar{x}, \mathbf{B}}; \theta_{ac}, \theta_{box})  \\
&= ReLU( \theta_z * \mathbf{\bar{z}}  +_b \theta_x * \mathbf{\bar{x}} +_b \eta(\mathbf{B}, \theta_{box}) ); \\
\end{aligned}
\end{equation}
Here, $B$ is the b-box of the initial frame and $\eta$ is a three-layer fully-connected network with parameters $\theta_{box}$. Since the target in the template is always at the center of the image, we only use the width and height of the b-box. Fig.~\ref{Fig:index_exp}(b) shows a tracking comparison between our ACM-based tracker and the baseline (using DW-XCorr) on example frames, where target is only part of an object (\eg, part of hand or body). 


To further demonstrate the effectiveness of our fusion, we conduct a simple experiment for digit prediction on MNIST dataset~\cite{lecun2010mnist}. 
First, we concatenate number images from MNIST into a $2 \times 2$ matrix and randomly generate an index of 0-3 to indicate the position of the numbers. Then, we design a network similar to AlexNet to predict the number at a given position. To incorporate the index information (a single number), we extract the index features using a three-layer fully-connected network and fuse them with the feature map of a matrix image using our ACM. We then feed the fused features into a prediction network. As shown in {Fig.~\ref{Fig:index_exp}(a)}, high responses are uniform without using index information. After integrating index information using ACM, they are more concentrated around the target positions. Even though we only give the network a single index number, it is able to better discriminate target position with emphasis to the region belonging to the target. As a result, our network correctly predicts the number at given position. 

\begin{figure}
  \centering
  \includegraphics[width = 8cm]{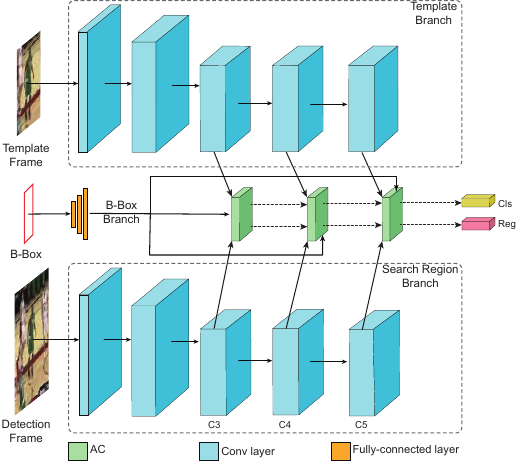}
  \caption{\textbf{Overview of our tracker (SiamBAN-ACM)} which integrates ACM, in place of DW-XCorr, in the baseline SiamBAN. 
  }
  \label{Fig:network}
\end{figure} 

\subsection{ACM for Visual Tracking}  \label{section:tracking}
The proposed ACM is generic and can be easily integrated into existing Siamese trackers. Here, we integrate ACM into three trackers: SiamFC~\cite{bertinetto2016fully}, the recently introduced SiamRPN++~\cite{li2019siamrpn} and SiamBAN~\cite{chen2020siamese}. For SiamFC, we replace its XCorr with ACM, whereas for SiamRPN++ and SiamBAN we replace their DW-XCorr with ACM. The resulting ACM-based trackers are named, SiamFC-ACM, SiamRPN++ACM and SiamBAN-ACM, respectively.


\noindent\textbf{{Our SiamFC-ACM.}} The original SiamFC~\cite{bertinetto2016fully}
employs XCorr to produce a single-channel response map. We use the same network as the original SiamFC to extract features, and feed the feature maps produced by the template and search region branches into ACM, producing a correlation map with a single channel. The position with the highest response is then set as the predicted target center.



\noindent\textbf{{Our SiamRPN++ACM.}} The original SiamRPN++~\cite{li2019siamrpn} is the first to introduce DW-XCorr into Siamese trackers. For SiamRPN++ACM, we replace the DW-XCorr in the original SiamRPN++ with our ACM. Specifically, ACM fuses the features from the three branches (template, search region and b-box) to generate a correlation feature map, as shown in Fig.~\ref{Fig:network}. The b-box branch uses three fully-connected layers to generate a target location feature map ($1 \times 1 \times 256$). Then, we apply two $5 \times 5$ convolutions without padding  to the template and search region feature maps to obtain semantic feature maps. Consequently, the summation of the three feature maps (\textit{i.e.} template, search region and b-box maps) is then batch normalized and used as input to the RPN heads. The template and initial b-box are fixed during inference and the three branches remain independent until the broadcasting summation. Thus, we can cache the two branches (template and b-box) to save computational cost. In this way, the additional computational cost introduced by ACM is only a single convolution on the search region, thereby causing no significant degradation to the overall inference speed. 


\noindent\textbf{{Our SiamBAN-ACM.}} The recent  SiamBAN~\cite{chen2020siamese} does not employ pre-defined anchors, enabling it to perform better and faster than its baseline SiamRPN++. To obtain SiamBAN-ACM, we apply same changes (replacing DW-XCorr with ACM) to the baseline SiamBAN as described above for SiamRPN++ACM.

\section{Experiments}
We perform comprehensive experiments on six tracking benchmarks: OTB-100~\cite{wu2015object}, UAV123~\cite{mueller2016a}, TrackingNet~\cite{muller2018trackingnet}, VOT2016, VOT2019~\cite{VOT_TPAMI} and LaSOT~\cite{fan2019lasot}. 
A well-documented and complete training and inference code will be publicly released.




\noindent\textbf{{Implementation Details.}} 
Our ACM-based tracking frameworks are implemented using the Pytorch tracking platform PySOT. For fair comparison, we follow the same training protocol (datasets and training hyper-parameters) for our SiamFC-ACM, SiamRPN++ACM and SiamBAN-ACM as that of their respective baseline SiamFC, SiamRPN++ and SiamBAN trackers. Further, we use the same loss functions in our tracking networks as that of the respective baselines, as ACM can be optimized without auxiliary guidance. We perform training on a workstation with an Intel E5-2698 v4 CPU, 512G memory, and four V100 GPUs. For both training and testing, template patches are cropped to $127\times 127$ pixels, and the search region is cropped to $255 \times 255$ pixels.

\subsection{ State-of-the-Art Comparison}
\noindent\textbf{TrackingNet~\cite{muller2018trackingnet}:} Table~\ref{Table:trackingnet} shows the comparison on TrackingNet test set, which comprises over 500 videos without publicly available
ground-truths. The results are obtained through an online evaluation server. Our three trackers (SiamFC-ACM, SiamRPN++ACM and SiamBAN-ACM) achieve consistent improvement over their respective baselines (SiamFC, SiamRPN++ and SiamBAN). The recently introduced KYS~\cite{bhat2020know} and its baseline DiMP~\cite{bhat2019learning} achieve normalized precision (\textbf{NP}) scores of 80.0 and 80.1, respectively.~Our SiamBAN-ACM achieves \textbf{NP} score of 81.0, outperforming both KYS and DiMP. SiamBAN-ACM also achieves favorable result in terms of success (\textbf{A}), against existing trackers with an AUC score of 75.3.

\begin{figure}[t] 
  \centering
  \includegraphics[width = 8cm]{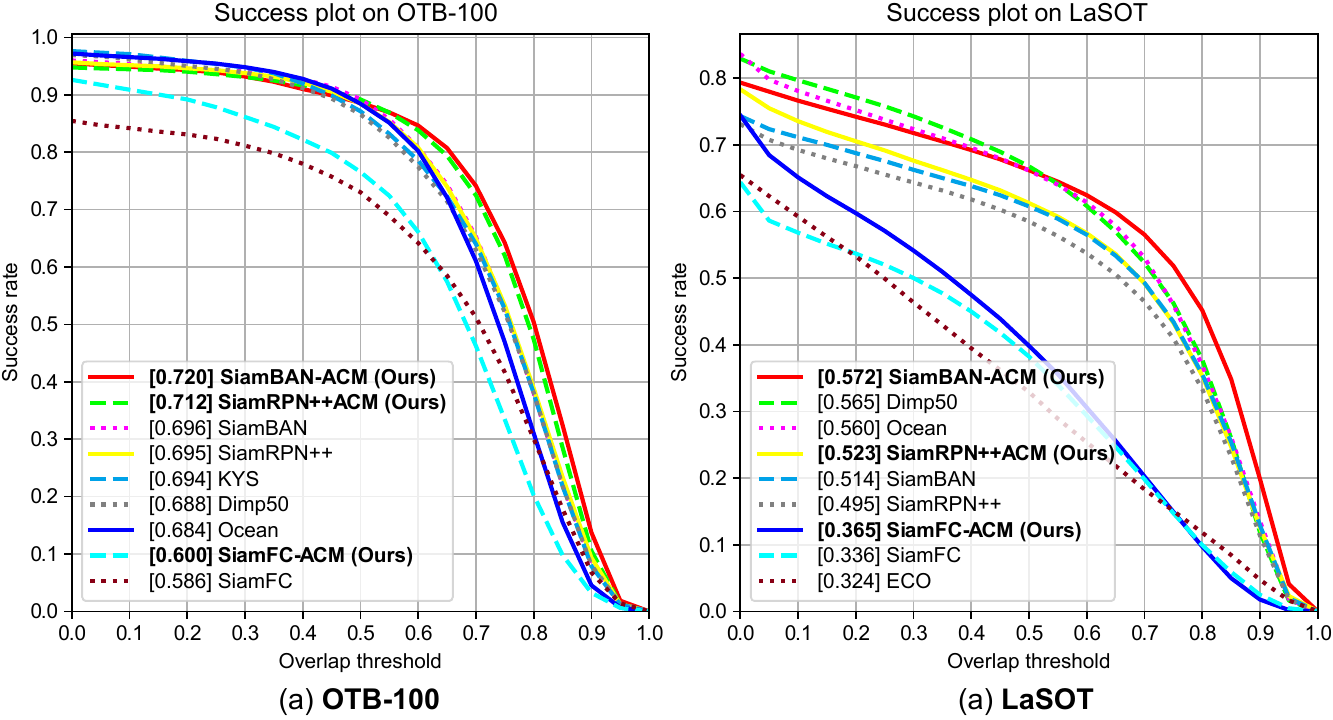}
  \caption  {\textbf{State-of-the-art comparison on (a) OTB-100~\cite{wu2015object} and (b) LaSOT~\cite{fan2019lasot} test set} in terms of success plot. For each method, we show the AUC scores in the legend. On both datasets, our ACM-based trackers (SiamFC-ACM,  SiamRPN++ACM  and  SiamBAN-ACM)  consistently outperform their respective baselines (SiamFC, SiamRPN++ andSiamBAN).  Best viewed zoomed in.}  
  \label{Fig:otb}
\end{figure}

\noindent\textbf{OTB-100~\cite{wu2015object}:}  Fig.\ref{Fig:otb}(a) shows the results, in terms of success plot, over all 100 videos of OTB-100. The trackers are ranked in terms of their AUC score (in the legend).
Among existing methods, SiamBAN achieves an AUC score of 69.6. The recently introduced KYS~\cite{bhat2020know} and its baseline DiMP~\cite{bhat2019learning} obtain AUC scores of 69.4 and 68.8, respectively. Our SiamBAN-ACM outperforms existing trackers with an AUC score of 72.0. Further, our SiamBAN-ACM obtains an absolute gain of 2.5\% over the baseline SiamBAN. In OTB-100, each video is annotated with 11
different attributes. SiamBAN-ACM achieves promising performance on all these attributes, compared to existing methods. The attribute plots are provided in the supplementary material. 

\begin{table}[t]
\fontsize{7}{8}\selectfont 
\centering
\setlength{\tabcolsep}{0.1mm}{
\begin{tabular}{cccccccccccc}
\hline
  & SiamFC ~& \textbf{SiamFC}~~& DiMP ~& SiamRPN++  ~& SiamBAN ~& KYS ~& \textbf{SiamRPN++}~& \textbf{SiamBAN}\\
 & \cite{tao2016siamese}& \textbf{-ACM} &\cite{bhat2019learning}&\cite{li2019siamrpn}& \cite{chen2020siamese} & \cite{bhat2020know} &\textbf{ACM}&\textbf{-ACM}\\
\hline
\textbf{A}  & 0.571 & 0.577 &   0.740 & 0.733 &   0.725~ &0.740 & {\color{blue}0.747} &{\color{red}0.753}\\
\hline
\textbf{P} & 0.553 & 0.537  &  0.687 & 0.694 &  0.687~ &0.688& {\color{blue}0.705}& {\color{red}0.712}\\
\hline
\textbf{NP} & 0.652 & 0.675  &  0.801 & 0.800 &  0.795~ &0.800& {\color{blue}0.804}& {\color{red}0.810}\\
\hline
\end{tabular}}
\caption{\textbf{State-of-the-art comparison on TrackingNet~\cite{muller2018trackingnet} test set} in terms of success (AUC), precision and normalized precision. Success, precision and normalized precision are denoted by \textbf{A}, \textbf{P} and \textbf{NP}, respectively. The best two results are shown in red and blue, respectively. 
} 
\label{Table:trackingnet}
\end{table}

\begin{table}[t]
\fontsize{6.7}{8}\selectfont
\centering
\setlength{\tabcolsep}{0.1mm}{
\begin{tabular}{cccccccccc}
\hline
&SiamFC~~& \textbf{SiamFC} ~~& SiamRPN++& ROAM++~~&SPM ~~& \textbf{SiamRPN++}~~& SiamBAN~~& \textbf{SiamBAN}\\
&\cite{bertinetto2016fully}&\textbf{-ACM}&\cite{li2019siamrpn}&\cite{yang2020roam}&\cite{wang2019spm}&\textbf{ACM}&\cite{chen2020siamese}&\textbf{-ACM}\\
\hline
\textbf{E}&0.277 &  { 0.338} &  0.441 & 0.434&0.481 &  0.501 &{\color{blue}0.505}&{\color{red}0.549}\\
\hline
\textbf{R}&0.382 &  { 0.294} &  0.174 & 0.210&0.206 &  {\color{blue}0.144} &0.149&{\color{red}0.098}\\
\hline
\textbf{A}&0.549 &  { 0.535} & 0.599 & 0.620&0.610 &  {\color{red}0.666} &0.632&{\color{blue}0.647}\\
\hline
\end{tabular}}
\caption{\textbf{State-of-the-art comparison on VOT2016 challenge dataset~\cite{VOT_TPAMI}} in terms
of expected average overlap (\textbf{E}), robustness (\textbf{R}) and accuracy (\textbf{A}).
The  best  two  results  are  shown  in red  and  blue  fonts, respectively.  
}
\label{Table:vot2016}
\end{table}

\begin{table}[t]
\fontsize{6.8}{8}\selectfont
\centering
\setlength{\tabcolsep}{0.1mm}{
\begin{tabular}{cccccccccc}
\hline
&SiamFC~~& \textbf{SiamFC} ~~& SiamRPN++~~& \textbf{SiamRPN++}~~& DiMP~~& SiamBAN~~& Ocean ~~& \textbf{SiamBAN}\\
&\cite{bertinetto2016fully}&\textbf{-ACM}&\cite{li2019siamrpn}&\textbf{ACM}&\cite{bhat2019learning}&\cite{chen2020siamese}&\cite{zhang2020ocean}&\textbf{-ACM}\\
\hline
\textbf{E}&0.163 &  { 0.206} &  0.285 & 0.303&0.321  &0.327&  {\color{blue}0.350} &{\color{red}0.362}\\
\hline
\textbf{R}&0.958 &  { 0.712} &  0.482 & 0.431&0.371  &0.396&  {\color{red}0.316} &{\color{red} 0.316}\\
\hline
\textbf{A}&0.470 &  { 0.503} & 0.599 &{\color{red}0.624}&0.582 &0.602& 0.594 &{\color{blue}0.621}\\
\hline
\end{tabular}}
\caption{\textbf{State-of-the-art comparison on VOT2019 challenge dataset~\cite{VOT_TPAMI}} in terms
of expected average overlap (\textbf{E}), robustness (\textbf{R}) and accuracy (\textbf{A}).
The  best  two  results  are  shown  in red  and  blue  fonts, respectively. 
}
\label{Table:vot2019}
\end{table}

\begin{table}[t]
\fontsize{6.6}{8}\selectfont
\centering
\setlength{\tabcolsep}{0.1mm}{
\begin{tabular}{cccccccccc}
\hline
SiamFC~~& \textbf{SiamFC} ~~& SiamRPN++ ~~& DiMP ~~& \textbf{SiamRPN++}~~& SiamCAR~~& SiamBAN~~& \textbf{SiamBAN}\\
\cite{bertinetto2016fully}&\textbf{-ACM}&\cite{li2019siamrpn}&\cite{bhat2019learning}&\textbf{ACM}&\cite{guo2020siamcar}&\cite{chen2020siamese}&\textbf{-ACM}\\
\hline
0.498 &  { 0.508}  &  0.613 &  {\color{red}0.654} & 0.634&0.614&0.631&{\color{blue}0.648}\\
\hline
\end{tabular}}
\caption{\textbf{State-of-the-art comparison on UAV123~\cite{mueller2016a}} in terms of success (AUC). The  best  two  results  are  shown  in red  and  blue  fonts,  respectively. } 
\label{Table:uav}
\end{table}

\noindent\textbf{LaSOT~\cite{fan2019lasot}:}
We evaluate our approach on the test set comprising 280 long videos. Fig.\ref{Fig:otb}(b) shows the success plot. We rank the trackers w.r.t. their AUC scores (in the legend). Among existing methods, SiamBAN and DiMP obtain AUC scores of 51.4 and 56.5, respectively. Our SiamBAN-ACM obtains favorable results against the state-of-the-art, while outperforming baseline SiamBAN by an AUC gain of 5\%. 



\noindent\textbf{VOT 2016 and 2019~\cite{VOT_TPAMI}:} 
Table~\ref{Table:vot2016} and \ref{Table:vot2019} show a comparison on VOT 2016 and 2019, respectively. On VOT2016, our SiamBAN-ACM outperforms the previous best SiamBAN with a EAO (\textbf{E}) absolute gain of 4.4\%.
Similarly on VOT 2019, our three trackers (in bold) achieve consistent improvement in performance over their baselines. Compared to SiamBAN, our SiamBAN-ACM has 20\% lower failure rate, while also achieving improved tracking accuracy.  


%
\noindent\textbf{UAV123~\cite{mueller2016a}:}
Table~\ref{Table:uav} shows the comparison in terms of success (AUC). Among existing Siamese trackers, SiamCAR and SiamBAN achieve AUC scores of 61.4 and 63.1, respectively. Our SiamBAN-ACM achieves favorable performance against existing trackers with AUC score of 64.8.

\subsection{Ablation Study}
We perform an ablation study to analyze the impact of ACM in the
three tracking architectures. As discussed earlier, our ACM addresses the limitations of XCorr and DW-XCorr by introducing an asymmetric convolution (AC). Further, ACM also possesses the flexibility to incorporate additional (non-visual) information in the form of b-box size. Here, we also analyze the impact of only replacing  the XCorr or DW-XCorr with AC and not incorporating additional (b-box size) prior information. 
We perform ablation experiments on the VOT2016 and OTB-100 datasets. We follow the standard evaluation protocols of the respective datasets. On VOT2016, trackers are evaluated using expected average overlap
(EAO) score. The EAO score takes into account both robustness and accuracy. Here, robustness represents number of tracking failures, while accuracy indicates the average overlap between the ground-truth b-box and tracker prediction. On OTB-100, trackers are evaluated using the area-under-the-curve (AUC), which is obtained by averaging the overlap precision (OP) scores over a range of thresholds [0, 1]. Here, OP metric indicates the percentage of frames where intersection-over-union (IoU)
overlap between the ground-truth b-box and predictions from the tracker is greater than a certain threshold. 

Table~\ref{Table:ablation} shows the results using three baseline tracking architectures on both datasets. We also report the speed in terms of frames per second (FPS). Note that all speeds are reported on a GTX1080Ti GPU. On VOT2016, the baseline SiamBAN and SiamRPN++ achieve EAO scores of 50.5 and 46.4, respectively. A consistent improvement in tracking performance is obtained when replacing the DW-XCorr with our AC in these two baseline architectures. Our final ACM, which contains both the AC and the prior b-box size information, achieves significant improvement in performance over both the baselines. Our ACM-based trackers (SiamBAN-ACM and SiamRPN++ACM) obtain absolute gains of 4.4\% and 3.7\%, in terms of EAO, over their respective SiamBAN and SiamRPN++ baselines. In case of the baseline SiamFC, our ACM contains only the AC and no additional (non-visual) information, since SiamFC only needs to predict the center of the target. Our ACM-based tracker (SiamFC-ACM) obtains a significant gain of 6.1\% over the baseline SiamFC. Similarly, our ACM-based trackers also provide consistent improvements in tracking performance on their respective baselines on OTB-100. 

\begin{table}[t] 
\footnotesize
\centering
\setlength{\tabcolsep}{2.0mm}{
\begin{tabular}{c|cc|ccc}
\hline
&AC & ACM & VOT2016 & OTB2015 & Speed\\
& &  & (EAO) & (AUC Score) & (fps)\\
\hline
& &  & 0.505& 0.695 & 48 \\
SiamBAN& \checkmark&  & 0.535& 0.715 & 41 \\
& \checkmark&  \checkmark& \textbf{0.549}& \textbf{0.720} & 41 \\
\hline
& &  & 0.464& 0.695 & 46 \\
SiamRPN++& \checkmark&  & 0.485& 0.705 & 40 \\
& \checkmark&  \checkmark& \textbf{0.501}& \textbf{0.712} & 40 \\
\hline
& &  & 0.277& 0.586 & 190 \\
SiamFC& \checkmark&  & \textbf{0.338}& \textbf{0.600} & 172 \\
\hline
\end{tabular}}
\caption{\textbf{Ablation study on VOT2016~\cite{kristan2016the} and OTB-100\cite{wu2015object}}. We show the results using three different baseline tracking architectures. All speeds are reported on a GTX1080Ti GPU. We also show our ACM with only AC and without the integration of prior non-visual information. In all cases, our final ACM achieves consistent improvement in tracking performance over the baseline architectures. The best scores are highlighted in bold in each case.
}
\label{Table:ablation}
\end{table}

\section{Conclusion}
We propose a learnable module, called the asymmetric convolution (ACM), to efficiently fuse feature maps of different sizes in Siamese trackers. Our ACM addresses the limitations of standard DW-XCorr and benefits from large-scale offline training. Further, ACM possesses the flexibility to integrate useful non-visual information, such as the location (b-box size) of target b-box in the initial frame. We integrate ACM into three Siamese tracking architectures. Experiments on six datasets demonstrate that ACM-based trackers provide consistent improvement over their baselines, leading to favorable results against existing methods.  Also we believe ACM would benefit other tasks.

\bibliographystyle{plain}
\bibliography{cite}

\clearpage

\end{document}